\documentclass{article}

    \PassOptionsToPackage{numbers, compress}{natbib}

\usepackage[preprint]{neurips_2022}

\usepackage{microtype}
\usepackage{graphicx}
\usepackage{subfigure}
\usepackage{hyperref}       
\usepackage{booktabs} 
\usepackage{amsfonts}       
\usepackage{nicefrac}       

\usepackage{lipsum}
\usepackage{xcolor}

\usepackage[space=true]{accsupp}
\newcommand{\copyablespace}{\BeginAccSupp{method=hex,unicode,ActualText=00A0}\ \EndAccSupp{}}
\usepackage{listings} 
\definecolor{codegreen}{rgb}{0,0.6,0}
\definecolor{codegray}{rgb}{0.5,0.5,0.5}
\definecolor{codepurple}{rgb}{0.58,0,0.82}
\definecolor{backcolour}{rgb}{0.95,0.95,0.92}
\lstdefinestyle{mystyle}{
    backgroundcolor=\color{backcolour},   
    commentstyle=\color{codegreen},
    keywordstyle=\color{magenta},
    numberstyle=\tiny\color{codegray},
    stringstyle=\color{codepurple},
    columns=fullflexible,
    basicstyle=\ttfamily\tiny,
    breakatwhitespace=false,         
    breaklines=true,                 
    captionpos=t,                    
    keepspaces=true,                 
    numbers=left,   
    showspaces=true,                
    showstringspaces=true,
    showtabs=true,                  
    tabsize=2,
    literate={\ }{{\copyablespace}}1
}
\usepackage{amsmath}

\usepackage{enumitem}
\title{Tuple Packing:\\
Efficient Batching of Small Graphs in 
Graph Neural Networks}

\author{%
Mario Michael Krell\\ 
  Graphcore Inc. \\
  United States of America\\
\And 
Manuel Lopez\\ 
  Graphcore Inc. \\
  United States of America\\ \And
Sreenidhi Anand\\ 
  Graphcore Inc. \\
  United States of America\\ \And
Hatem Helal\\
  Graphcore Inc. \\
  England\\ \And
Andrew William Fitzgibbon\\
  Graphcore Inc. \\
  England}

\begin{document}

\maketitle

\begin{abstract}
When processing a batch of graphs in machine learning models such as Graph Neural Networks (GNN),
it is common to combine several small graphs into one overall graph 
to accelerate processing and remove or reduce the overhead of padding.
This is for example supported in the PyG library.
However, the sizes of small graphs can vary substantially 
with respect to the number of nodes and edges, 
and hence the size of the combined graph can still vary considerably, 
especially for small batch sizes.
Therefore, 
the costs of excessive padding and wasted compute are still incurred
when working with static shapes, which are preferred for maximum acceleration.
This paper proposes a new hardware agnostic approach ---tuple packing--- 
for generating batches that cause minimal overhead.
The algorithm extends recently introduced sequence packing approaches to
work on the 2D tuples of $(|\text{nodes}|, |\text{edges}|)$.
A monotone heuristic is applied to the 2D histogram of tuple values  
to define a priority for packing histogram bins together
with the objective to reach a limit on the number of nodes as well as the number of edges.
Experiments verify the effectiveness of the algorithm
on multiple datasets.
\end{abstract}

\section{Introduction}

\begin{figure*}[htb!]
    \centering
    \includegraphics[width=0.32\linewidth]{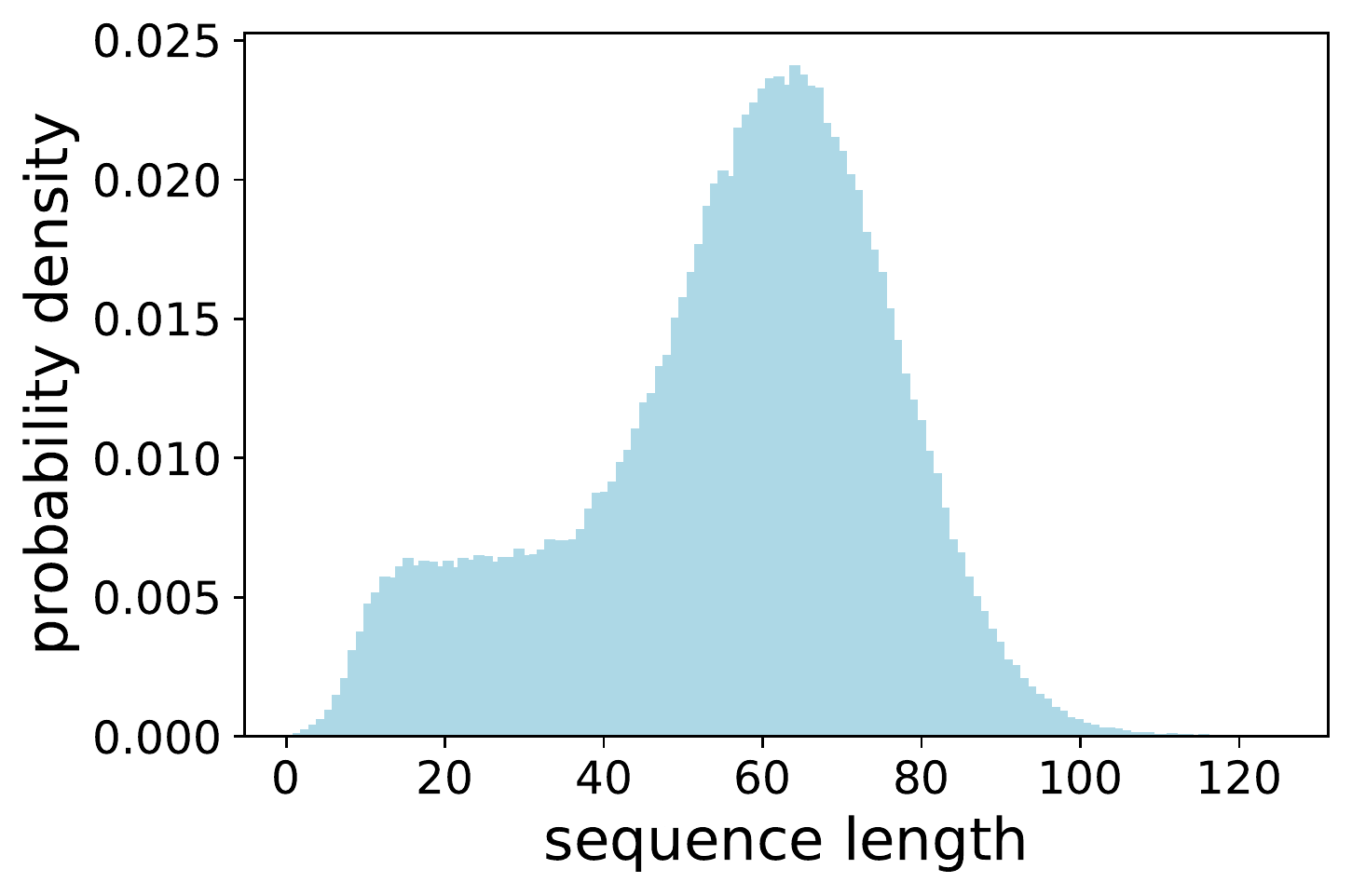}
    \includegraphics[width=0.32\linewidth]{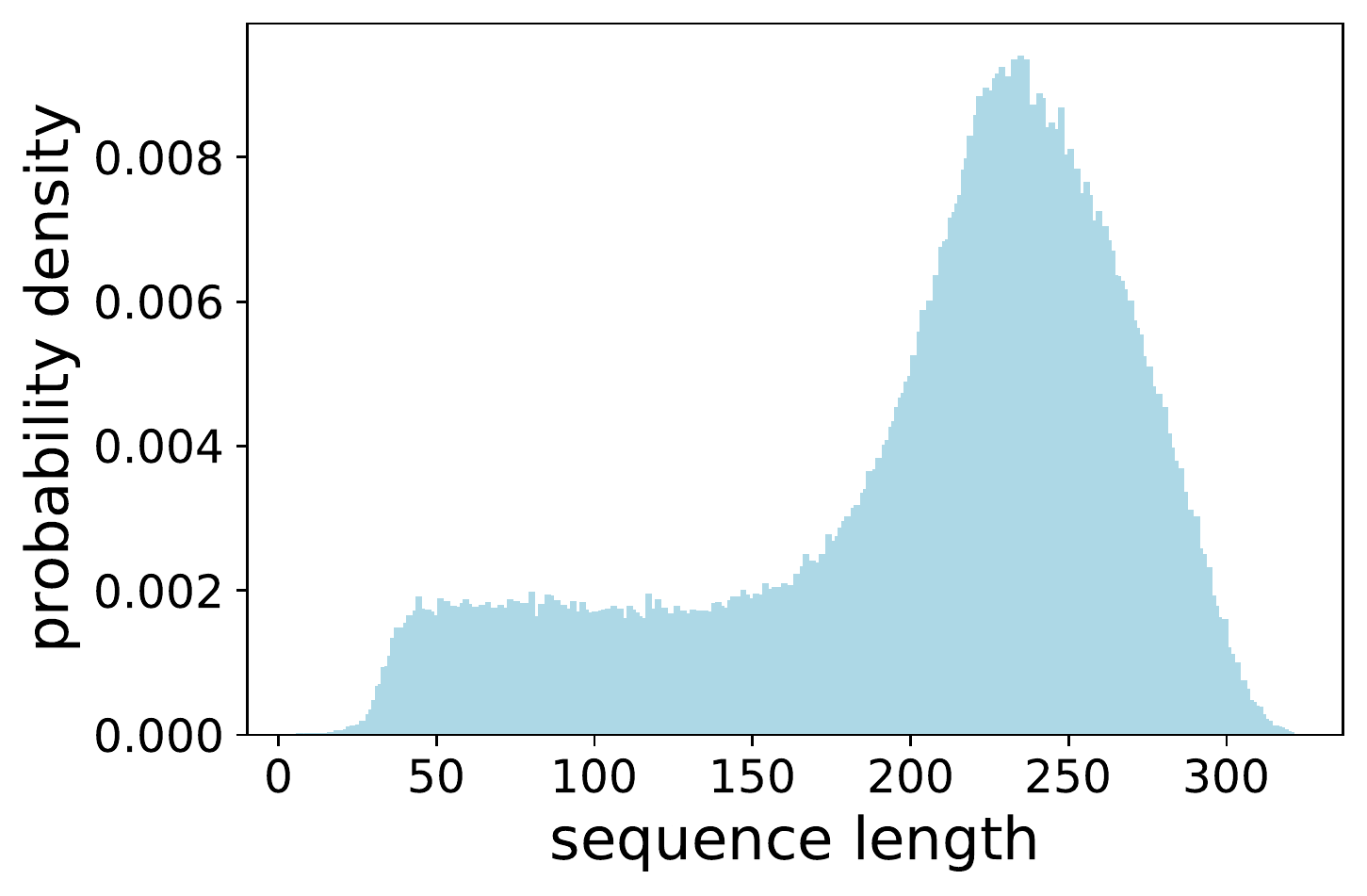}
    \includegraphics[width=0.32\linewidth]{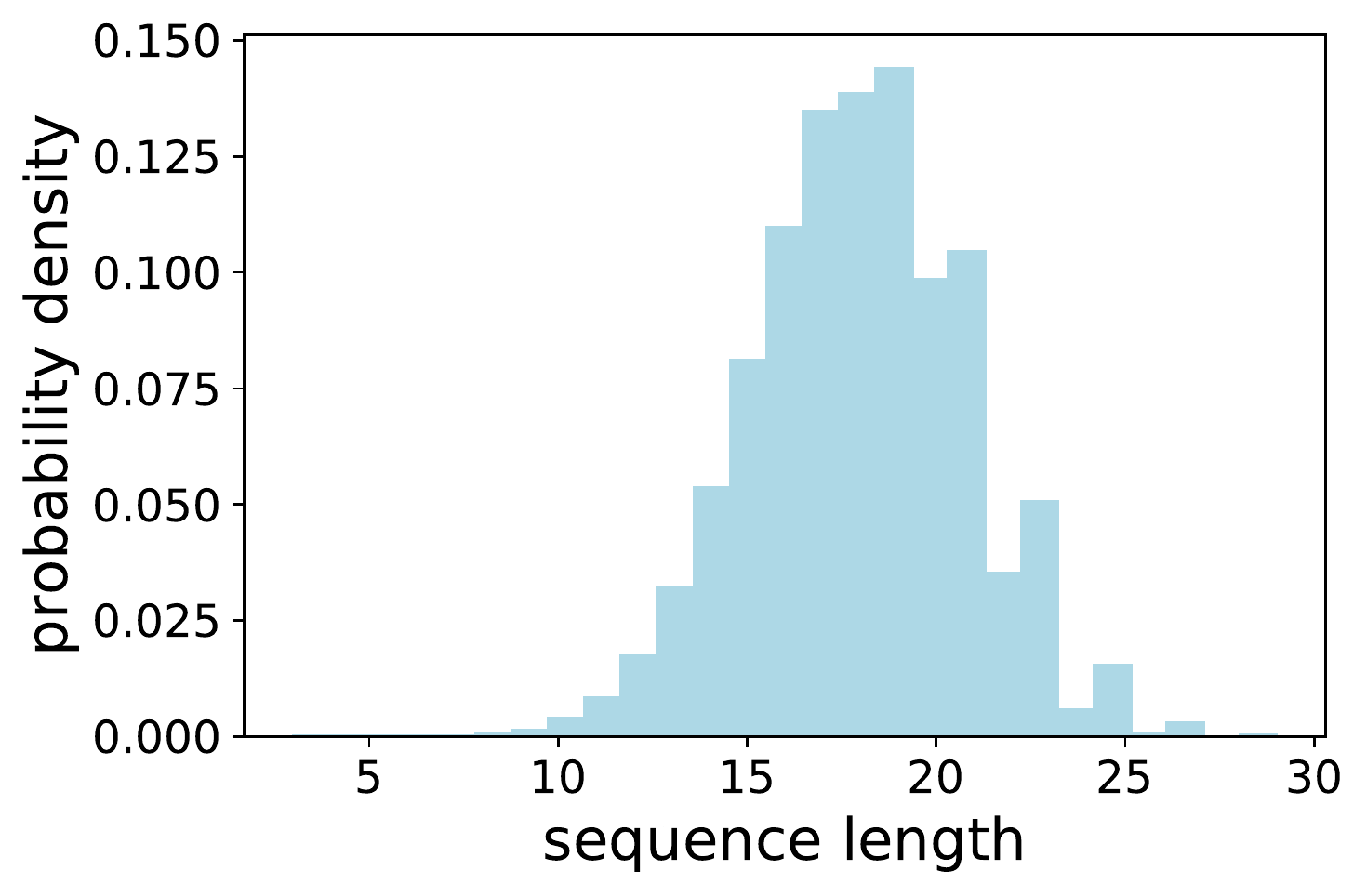}
    \caption{Length distributions for different datasets. 
    From left to right: 
    LibriSpeech text labels,
    LibriSpeech audio token sequence~\cite{panayotov2015librispeech} ,
    and QM9 molecules of a graph as a node sequence~\cite{qm9,ramakrishnan2014quantum} .
    }
    \label{f:datasets}
\end{figure*}

Small graphs can be obtained from a variety of data sources 
(for example text, audio, and molecules).
As shown in Figure~\ref{f:datasets}, the resulting length of data (number of nodes) can vary substantially.
The resulting padding can have a significant impact on transformer models like BERT~\cite{Devlin2019}
and can largely benefit from combining/packing data together~\cite{Kosec2021}.
The main advantage of the data packing technique is that it is hardware agnostic
and results in speedup on Intelligence Processing Units (IPUs) by Graphcore~\cite{Kosec2021}, 
GPUs by NVIDIA~\footnote{BERT sequence packing section at:\\ \url{https://developer.nvidia.com/blog/boosting-mlperf-training-performance-with-full-stack-optimization/}}, 
and Gaudi by Intel/Habana~\footnote{\url{https://developer.habana.ai/tutorials/tensorflow/data-packing-process-for-mlperf-bert/}}.

\begin{figure*}[htb!]
    \centering
    \includegraphics[width=0.49\linewidth]{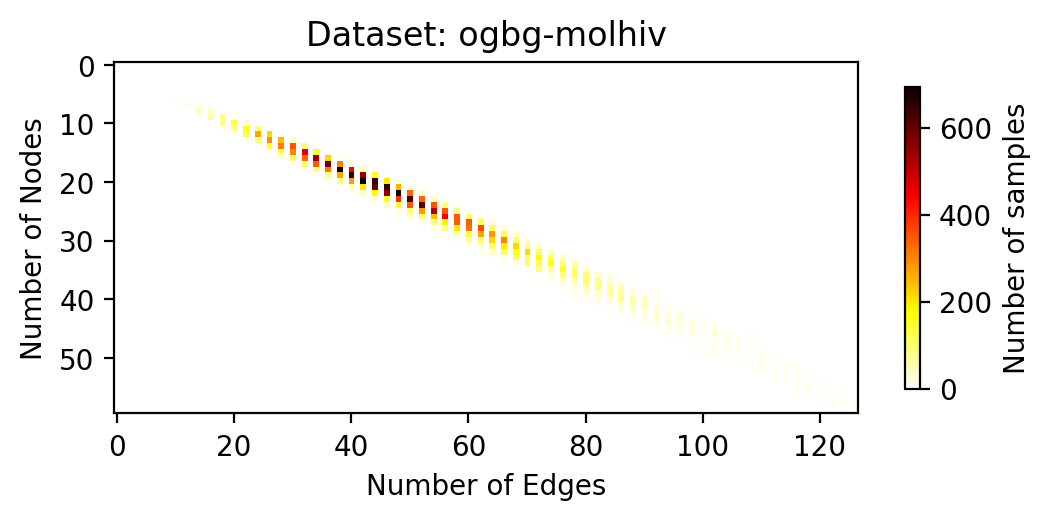}
    \includegraphics[width=0.49\linewidth]{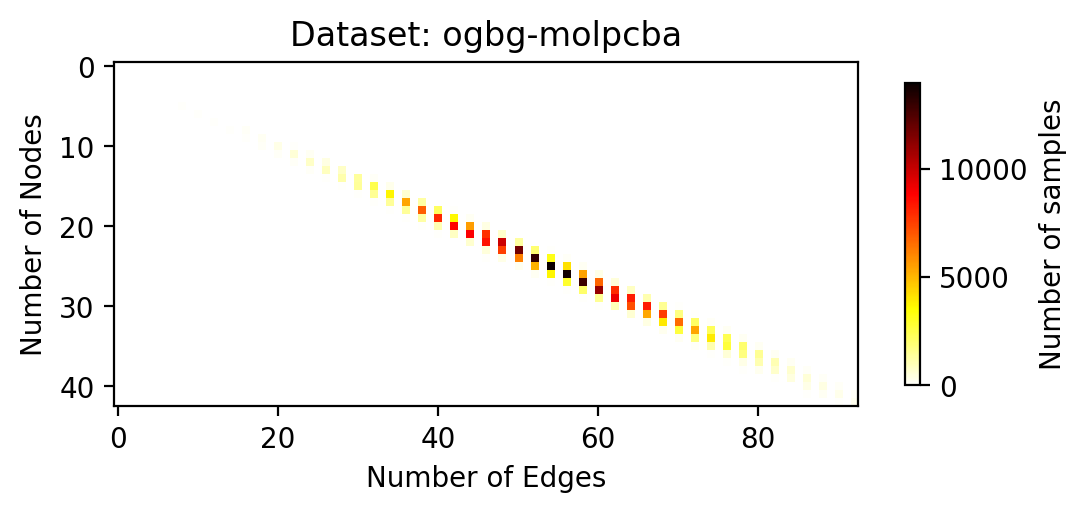}
    \includegraphics[width=0.49\linewidth]{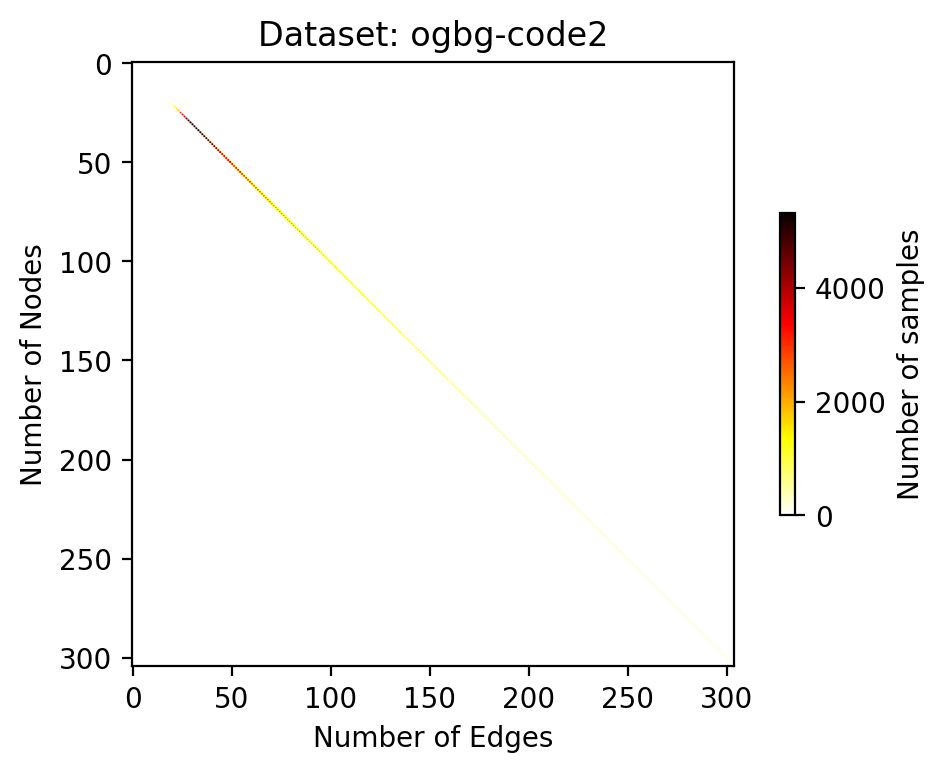}
    \includegraphics[width=0.49\linewidth]{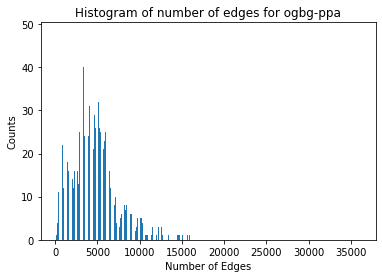}
    \caption{Histograms of other small graph OGB datasets for graph property prediction. 
        For the ogbg-ppa dataset, more than $50\%$ of all samples have $300$ nodes and
        counts are very small which prevents a proper 2D visualization.
    }
    \label{f:ogb_histogram}
\end{figure*}

\begin{figure*}[htb!]
    \centering
    \includegraphics[width=0.9\linewidth]{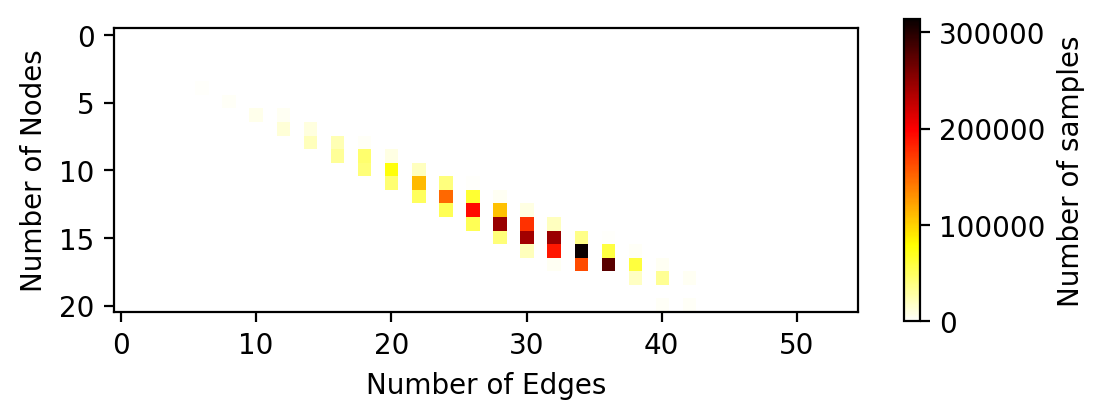}
    \includegraphics[width=0.49\linewidth]{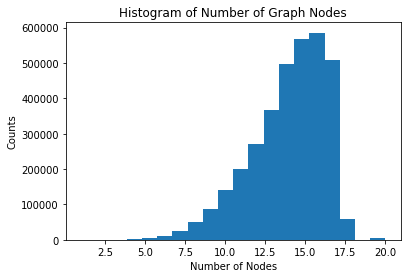}
    \includegraphics[width=0.49\linewidth]{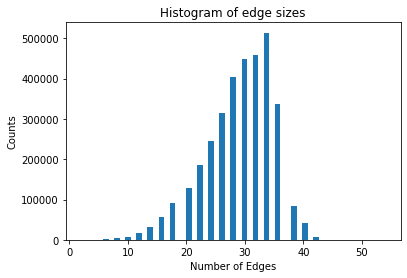}
    \caption{Histograms of the PCQM4Mv2 molecule graph dataset.
    For the number of edges, only multiples of two occur,
    since the graph is undirected but the underlying dataset description
    also allows for a directed graph.
    }
    \label{f:dual_histogram}
\end{figure*}

In this paper, we focus our application on Open Graph Benchmark (OGB) datasets for graph property 
prediction~\cite{hu2020ogb}~\footnote{\url{https://ogb.stanford.edu/docs/graphprop/}}
as well as the PCQM4Mv2 dataset~\cite{PubChemQC} that is part of the OGB large scale 
competition~\cite{hu2021ogblsc}~\footnote{\url{https://ogb.stanford.edu/docs/lsc/pcqm4mv2/}}.
The histograms are provided in Figure~\ref{f:ogb_histogram} 
and Figure~\ref{f:dual_histogram} respectively.
In some cases, the long tails of the distributions were cut off for better visualization.
Statistics are provided in Table~\ref{tab:datasets}.
The node-edge length count describes how many different combinations
of the number of edges and nodes exist in the dataset.
Further datasets and statistics are described on the PyG 
webpage~\footnote{\url{https://pytorch-geometric.readthedocs.io/en/latest/notes/data_cheatsheet.html}}.
We used PyG to obtain the data distributions for the experiments in this paper.

\begin{table}[ht!]
\caption{
Dataset properties of different graph datasets including padding efficiency.
}
\label{tab:datasets}
\begin{center}

\begin{tabular}{l|rrrrrrr}
\hline
            &  \multicolumn{4}{c}{ogbg-datasets}\\
Metric & molhiv & molpcba & code2 & pcqm4mv2 & ppa \\
\hline
Number of (training) graphs      & 32901 & 350343 & 407976 & 3378606 & 78200\\
Max number of nodes              & 222  & 313 & 36123 & 20 & 300 \\
Max number of edges              & 502  & 636 & 36122 & 54 & 36138\\
Node-edge length count           & 795  & 576 & 2099  & 152 & 35981\\
Node efficiency (\%)             & 11.4 & 8.21 & 0.35 & 27.7 & 81.1\\
Edge efficiency (\%)             & 10.8 & 8.71 & 0.34 & 24.7 & 12.6\\
Potential speedup for nodes      & 8.8  & 12.2& 288.8 & 3.6 & 1.23\\
Potential speedup for edges      & 9.3  & 11.5& 291.2 & 4.1 & 7.95\\
\hline
\end{tabular}
\end{center}
\end{table}

In all cases, a substantial amount of padding 
for the nodes and edges is required in order to obtain equally sized batches.
Apart from the PPA dataset, 
we can observe an almost linear relationship 
between the number of edges and the number of nodes.
Hence, two questions arise:
firstly, can we generalise the packing concept introduced for BERT~\cite{Kosec2021} to tuples?
This question is addressed in Section~\ref{s:graphmethods}.
Secondly, how much benefit would a tuple packing approach provide compared
to solely focusing the packing on edges or nodes and applying classical packing?
This is addressed in Section~\ref{s:graphexp}.
We summarize and provide an outlook in Section~\ref{s:graphconc}.

\section{Methods}
\label{s:graphmethods}

\subsection{Single Sequence Packing}

``\textit{The bin packing problem deals with the assignment of items into bins of a fixed capacity such that the number of utilized bins is minimized.
In the canonical formulation of the packing problem a vector $s(i)$ 
of length $n$ is used to represent the items being packed, where $s(i)$ denotes the length of the i-th sequence/item.
The allocation of items into bins is tracked through the assignment matrix $B$, where $B_{ij} \in \{0, 1\}$ states whether the i-th sequence should be
placed into the j-th bin. 
In the worst case scenario, every item is assigned to its own bin, thus $B \in \mathbb{R}^{n \times n}$.
Notably, $s$ grows linearly in the number of sequences/items being packed and $B$ grows with the square.
To mask out unused bins $y_j \in \{0, 1\}$, denotes whether the j-th bin is being used. 
The optimization objective is to minimize the sum of $y_j$
while making sure to assign each $s_i$ to exactly one bin and not exceeding the maximum bin capacity [constant] $s_m$ for each bin. [The index m stands for $\max$.]
This problem formulation is well known as bin packing~\cite{Korte2012}.}''\cite{Kosec2021} 

\begin{equation}
\label{e:base}
\begin{aligned}
\min_{y\in\{0,1\}^n,B\in\{0,1\}^{n\times n}} \quad & \sum_{j=1}^{n}{y_{j}} &
\text{Minimize the number of bins.} \\
\textrm{s.t.} \quad & \sum_{j=1} b_{ij} =1\quad   \forall i & 
\text{Assign each length/sequence to only one bin.}\\
  &\sum_{i=1}^n s(i)b_{ij}\leq s_m y_j \quad  \forall j & \text{Cumulative length cannot exceed capacity.}\\
\end{aligned}
\end{equation}

As discussed by Kosec et al.~\cite{Kosec2021}, 
this problem can be simplified by working on histograms instead.
It can be solved either by casting it to a non-negative least squares problem,
or by using simple heuristics to directly derive a solution 
(for example by applying first-fit decreasing or best-fit on the histograms).
Note that there are two heuristics present.
The first, more trivial one, decides how we sort the incoming sequences,
which is most efficient when going from longest to shortest.
The second heuristic decides how we measure the remaining space and 
how to sort bins that are not yet full.
Again, we can only use the sequence length of the remaining space in the bin.
However, first-fit sorts the remainder from longest to smallest and
best-fit uses the reverse order.
For tuples of sequences, this becomes more challenging.
More sophisticated heuristics are required 
to reduce the tuple of sequence lengths to a single number 
which is used to decide the packing priority.

\subsection{Tuple Packing}

Tuple packing means that instead of items with length (size) $s(i)$,
we have a tuple of lengths 
$$s(i)=(s^{(1)}(i),\ldots ,s^{(l)}(i))$$
for a l-tuple of items that need to be packed
up to a maximum capacity $s^{(k)}_m$ per component of the tuple:

\begin{equation}
\label{e:tuple}
\begin{aligned}
\min_{y\in\{0,1\}^n,B\in\{0,1\}^{n\times n}} \quad & \sum_{j=1}^{n}{y_{j}} &
\text{Minimize the number of bins.} \\
\textrm{s.t.} \quad & \sum_{j=1} b_{ij} =1\quad   \forall i & 
\text{Assign each length/sequence to only one bin.}\\
  &\sum_{i=1}^n s^{(k)}(i)b_{ij}\leq s^{(k)}_m y_j \quad  \forall j \forall k & \text{Cumulative length cannot exceed capacity.}\\
\end{aligned}
\end{equation}

\begin{figure}[ht!]
    \centering
    \includegraphics[width=0.8\linewidth]{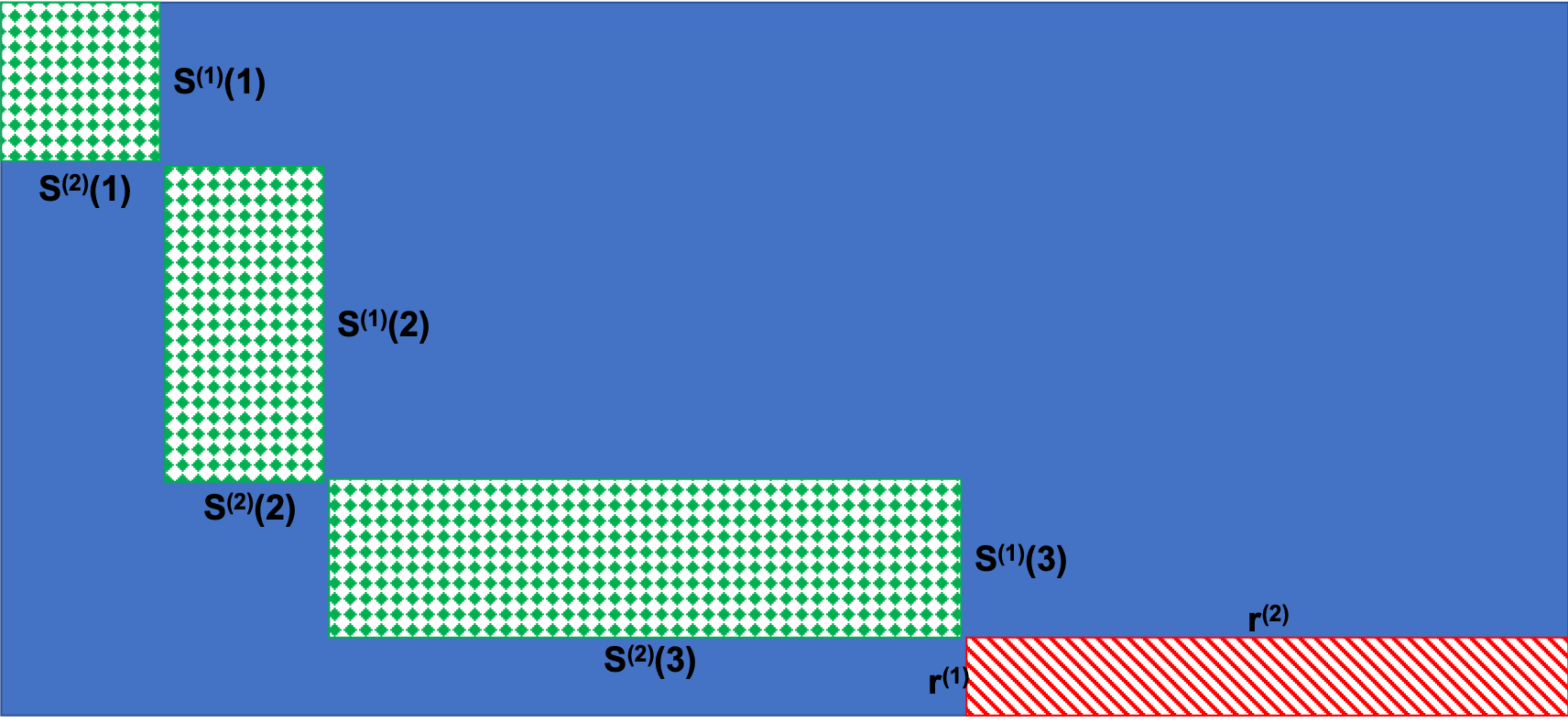}
    \caption{Visualization of the pair packing problem for a single bin.
    The remainder for the two components that require padding is denoted
    by $(r^{(1)},r^{(2)})$. 
    The objective is to minimize those two values.
    In contrast to 2D packing, 
    the blue area is not relevant for the optimal solution.
    }
    \label{f:recpack}
\end{figure}

Note that the number of parameters does not change but the number of constraints does.
Thus the higher the $k$, the more challenging it is to find a good solution.
Also note that when considering a vector/matrix notation
for the inequality in Equations~\ref{e:tuple},
it is equivalent to Equation~\ref{e:base}
Tuple packing is simpler than k-dimensional packing as shown in Figure~\ref{f:recpack}
where $k=2$.
Instead of filling a full rectangle, we are only packing along the ``diagonal''
and the remaining lower rectangle has to be minimized.

In practice, we are less interested in minimizing the number of bins but
we want the sum of
$$r^{(k)}_j:=s^{(k)}_m y_j-\sum_{i=1}^n s^{(k)}(i)b_{ij}$$ 
to be as small as possible for each $k$.
In other words, we want to have as little padding as possible
in order to reach the maximum capacity $s^{(k)}_m$ in each tuple component.
Note, however, that solving the optimization problem will provide the same result for both objectives
because the fewer bins we have, the less padding we encounter.

\subsection{Heuristics}

For single sequence packing, the objective is clearly to minimize
the single padding value, however the existence of a packing tuple $r^{(k)}_j$
for each bin $j$ requires some compromises.
Our proposal is to use a monotonically increasing heuristic $h$
that is zero when there is no remainder.
Examples are
$\min, \max, \prod, \sum$ and their weighted counterparts.
For pair packing, monotonically increasing means that
if $a\geq b$ and $c\geq d$ it holds $h(a,c)\geq h(b,d)$.
Strictly monotone means that 
if $a\geq b$ and $c > d$ (or $a> b$ and $c \geq d$) it holds 
$h(a,c) > h(b,d)$.
The functions $\min$ and $\max$ do not fulfill the strict condition
but are monotone.
The product $\prod$ is strictly monotone as long as no factor is zero.
The sum $\sum$ is strictly monotone.
Also the projection onto one component is monotone
and would reduce the problem to the well known bin-packing
if no limit for the other components is set.

\subsection{Tuple longest-pack-first histogram-packing}

Together with the heuristic, we propose a new variant of 
Best-Fit Bin Packing~\cite{johnson1973near,Dosa2014} applied
to a histogram in descending order 
(also known as longest-pack-first histogram-packing).

When packing simple sequences, we can sort by the sequence length,
however in this example, we need the heuristics.
Hence, we sort the incoming bins of node-edge pairs and respective counts based on heuristics.
This means that multiple tuples can obtain the same heuristic and
sorting might be not deterministic.
For example, for the $\prod$ heuristic (as well as the unweighted heuristics $\min, \max, \sum$ ), 
the pairs $(42,21)$ and $(21,41)$ get the same heuristic value of $861$.
Originally, for Best-Fit, we chose the fullest existing pack that could still fit the
incoming bin.
The term ``pack'' is used here to refer to a set of length tuples 
that were combined and their respective count of occurrence.
This time, each existing pack comes with a remainder in each component, 
that can still be filled (the tuple of $r^{(k)}_j$ described above).
We apply the same heuristic to each remainder of existing packs.
Next, we iterate from the bins starting with the value of the heuristic of the incoming bin
up to the maximum value of the heuristic applied to the length limits $s^{(k)}_m$.
For each value, we iterate over all respective existing packs and check
separate if the bin fits in.
For the single sequence algorithms, this check was always true and thus not required.
If the bin fits, it gets added to the pack and its respective count gets split and reduced.
If the count is not zero, the search starts anew from the beginning,
since the modification of the pack might have created a new pack that can fit
the incoming bin multiple times.
If the bin cannot fit in any pack, a new pack is created that includes only the bin.

It is also possible to implement a First-Fit variant 
(also called tuple shortest-pack-first histogram-packing).
The algorithm is exactly the same but starts with the packs with the largest value of the heuristic
down to the smallest, because the ``shortest'' pack will have the largest remainder.

Since the algorithms iterate on histograms instead of each item,
they can return a packing proposal within milliseconds if the number of bins is small.
This is crucial to iterate over different limits for $s^{(k)}_m$.
Compared to the sequence packing of the 
original algorithm~\cite{Kosec2021}, the complexity of tuple packing is 
at least as high (the number of sequences squared) 
with the maximum sequence length now replaced by the 
number of possible combinations of edges and nodes.
The number of possible combinations 
can be considerably higher compared to sequence packing in BERT.
A more detailed complexity analysis is planned to investigate this.

\subsection{Choice of length limits}
\label{s:limits}
For Natural Language Processing algorithms like BERT,
the maximum sequence length is fixed and cannot be adjusted for packing because
it would increase computing and memory requirements quadratically
and thus suppress any speedup from packing.
For GNNs, the cost is usually linear, and thus, the pack limits for tuples, $s^{(k)}_m$,
can be adjusted to requirements of the hardware as well as the packing efficiency.
So for example, instead of choosing $s^{(1)}_m$ 
as the maximum number of nodes of any graph in the dataset
and $s^{(2)}_m$ as the maximum number of edges, larger numbers are possible
where the limit is usually given either by the hardware or by the
processed batch becoming too large and updates happening too frequently.

Given that the packing algorithms provide a solution very quickly,
we propose to iterate over all potential combinations of relevant limits
from smallest to longest and choose the best trade-off by hand.
This could be, for example, the smallest combination (calculated by the heuristic)
that achieves a certain packing efficiency (say $95\%$) either on average or
for each component.
Note that starting with the maximums in the dataset and increasing
each of them step-wise by one is not recommended because
it will favor one component and suppress efficiency in the other components.

If the search space is too big, 
the limits can be linearly scaled until one components reaches high efficiency
and then the other components can be optimized.
It is also possible, to set no limit for all but one components to get a good start
and iterate from a solution that is optimal for one component.

It is future work to determine more automatic search algorithms.
Note that any derivative free parameter optimization algorithm can be applied,
for example direction search,
because a single evaluation can be calculated fast.

\subsection{Implementation}

Whilst transformers still require extra effort to be made to process packed samples,
it is becoming standard for Graph Neural Networks libraries 
to include the processing of packed samples.
For example in PyTorch Geometric (PyG), all the data that needs to be combined
is provided as a batch which gets automatically combined and processed as a large 
graph~\footnote{\url{https://pytorch-geometric.readthedocs.io/en/latest/notes/batching.html}}.
The PyG documentation states:
``\textit{This procedure has some crucial advantages over other batching procedures:
\begin{enumerate}
    \item 
    GNN operators that rely on a message passing scheme do not need to be modified since messages still cannot be exchanged between two nodes that belong to different graphs.
    \item
    There is no computational or memory overhead. For example, this batching procedure works completely without any padding of node or edge features. Note that there is no additional memory overhead for adjacency matrices since they are saved in a sparse fashion holding only non-zero entries, i.e., the edges.
\end{enumerate}
PyG automatically takes care of batching multiple graphs into a 
single giant graph with the help of the \texttt{torch\_geometric.loader.DataLoader} class.}''
Note that for obtaining static data shapes, some padding is still required. 
However, this can be kept to a minimum with a good packing strategy.

A similar approach exists in the JAX JGraph 
library\footnote{\url{https://jraph.readthedocs.io/en/latest/api.html\#batching-padding-utilities}}
and the TensorFlow DGL library~\cite{wang2019dgl}\footnote{\url{https://docs.dgl.ai/en/0.8.x/_modules/dgl/batch.html}}.
In all three libraries, there is a reverse operation
to separate the graphs again.
The two operations are sometimes 
called batching and unbatching. 
Unbatching is for example useful 
to add a transformer head\cite{graphgps2022}
or to calculate a graph-wise loss.
Note that the packing techniques
can still be applied with a modified unbatching.

Note that packing can limit the variation in the dataset -
the packs get shuffled but the graphs that are combined in a pack might stay fixed.
However, there will be still a large amount of variation in the bact.
It is also possible to only fix the sizes of graphs that get combined but sample the graphs randomly.
The benefit of this techniques largely depends on the dataset.
In an informal analysis, we did not observe any negative impact 
on performance when using packing and there was no noticeable
effect from randomizing the content of packs further.
This is aligned with the observations in BERT~\cite{Kosec2021}.
Further analysis however is required
for a more complex study of this technique.

\section{Experiments}
\label{s:graphexp}

This section evaluates different aspects 
of tuple packing such as the choice of heuristic and of length.
To evaluate the algorithm, the node and edge efficiencies 
have to either be provided separately 
or aggregated in a joint metric.

\subsection{Heuristic comparison}

In the first set of experiments, we used the maximum number of edges and nodes in the dataset as a limit for the packing
and compared different heuristics against baselines.
In the first four heuristics, the number 
of remaining edges and nodes get combined.
For the other two heuristics, we projected and only looked at the remaining spots for nodes or edges.
Note, that the limits for the nodes and edges are still applied.
For the baselines, we considered no packing (None)
or applying packing only on one component while totally ignoring the second component (Node base and Edge base).
Thus, the maximum number of edges or nodes can be exceeded for the component that is not considered.
The results are displayed in Table~\ref{tab:heuristics}.

\begin{table}[ht!]
\caption{
Packing efficiency of node and edge packing provided as performance number tuple for different heuristics and datasets. 
Maximum number of tuples in a pack is set to 256.
Processing longer than 10 minutes was stopped.
}
\label{tab:heuristics}
\begin{center}

\begin{tabular}{l|rrrrrrr}
\hline
            &  \multicolumn{4}{c}{ogbg-datasets}\\
 Heuristics & molhiv & molpcba & code2 & pcqm4mv2 & ppa\\
\hline
Product & (95.6, 90.5) & (89.6, 95.1) & (\underline{46.0}, \underline{45.6}) & (\underline{76.1}, \underline{58.0}) & -\\
Sum     & (97.5, 92.4) & (85.4, 90.6) & (\underline{46.0}, \underline{45.6}) & (\underline{76.1}, \underline{58.0}) & -\\
Maximum & (98.5, 93.3) & (\underline{92.6}, \underline{98.3}) & (\underline{46.0}, \underline{45.6}) & (\underline{76.1}, \underline{58.0}) & -\\
Minimum & (98.5, 93.3) & (\underline{92.6}, \underline{98.3}) & (\underline{46.0}, \underline{45.6}) & (\underline{76.1}, \underline{58.0}) & -\\
Node    & (\underline{98.8}, \underline{93.6}) & (91.5, 97.1) & (26.0, 25.8) & (75.1, 57.2) & (99.3, 15.4)\\
Edge    & (98.5, 93.3) & (90.6, 96.2) & (\underline{46.0}, \underline{45.6}) & (\underline{76.1}, \underline{58.0}) & -\\
\hline
Baseline\\
\hline
None     & (11.4, 10.8) & (8.21, 8.71) & (0.35, 0.34) & (27.7, 24.7) & (81.1, 12.6) \\
Node base& (98.7, 85.9) & (88.3, 84.5) & (26.0, 25.8) & (75.1, 48.3) & (99.3, 15.4)\\
Edge base& (85.8, 93.3) & (81.0, 88.1) & (40.1, 40.1) & (62.7, 76.4) & (32.7, 99.9)\\
\hline
\end{tabular}
\end{center}
\end{table}

We did not measure times. However, all the results for molhiv, molpcba, and pcqm4mv2 were obtained in milliseconds.
For code2, it took several seconds but less than ten minutes.
For ppa, multiple results could not be obtained within this time limit.
This is reasonable due to the complexity of the search space:
code2 has $2099$ original bins and up to 
$36122\cdot36123$ size combinations, 
whereas ppa has $35981$ original bins and $300\cdot36138$ potential combinations.

There is no clear favorite heuristic that works for all cases.
Tuple packing with the individually best heuristic always 
improves performance significantly compared to the baselines.
However, the original packing algorithms 
are already effective 
and the anticipated additional speedup gains 
from also using tuple packing are mostly around 1.15x.

For pcqm4mv2, results are rather low because of the distribution of the data
where the limits on the nodes and edges do not really allow for perfect patting
since the majority of graphs is around 15 edges and 30 nodes,
whereas our maximum was 20 nodes and 54 edges.
Thus setting different limits is of interest.
Further investigation with the code2 dataset is required to get 
a faster packing algorithm 
that can handle combining more graphs 
with better efficiency. 
Note that most graphs in code2 have around $50$ nodes 
and $50$ edges whereas the maximum is around $36122$
nodes and edges.
This is a very imbalanced packing scenario where 
a lot of graphs need to be combined to avoid padding,
which explains the comparably low efficiency values.





\subsection{Size limit variation}

\begin{figure}[htb!]
    \centering
    \includegraphics[width=0.95\linewidth]{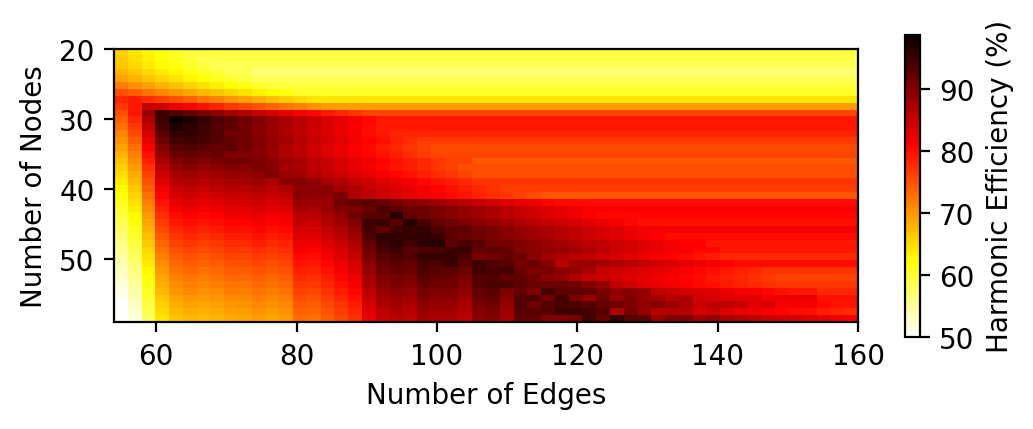}
    \caption{Harmonic mean of edge and node packing efficiency on PCQM4Mv2 
    for different maximum numbers
    of nodes and edges with optimum at 30 nodes and 62 edges at $(98.6\%, 99.0\%)$ efficiency, 
    closely followed by 45 nodes and 96 edges at $(94.9\%, 98.4\%)$).
    All small values were rounded up to $50\%$ for visualisation purposes.
    }
    \label{f:PCQM4Mv2results}
\end{figure}

For the PCQM4Mv2 dataset, we observed low efficiency of tuple packing 
on the edge component ($58\%$).
In this Section, we evaluate the method from Section~\ref{s:limits}
and compare different limits for the number of nodes and edges in a pack.
We use the product heuristic on the PCQM4Mv2 dataset 
with at least $20$ nodes and $54$ edges. 
The results are visualized in Figure~\ref{f:PCQM4Mv2results} with the baseline
in the upper left corner.
Obtaining these results only takes a short amount of time 
because of the fast packing algorithm.
It can be seen that the efficiency can be significantly improved.
For visualization purposes we have used the harmonic mean 
because it is sensitive to the balance of the two constituent efficiencies.
The best result was obtained for $30$ nodes and $62$ edges ($98.6\%, 99.0\%$)
with $98.8\%$ harmonic mean.
This is a $50\%$ improvement compared to the previous best result of $65.8\%$
as the harmonic mean of $(76.1, 58.0)$.

\section{Conclusion}
\label{s:graphconc}

In this paper we discuss how a large potion of padding potentially may be required for
processing graphs by visualizing multiple datasets.
To address this, we introduce a new tuple packing problem and solution.
We show that while normal packing 
(that addresses only one component of the input data)
can reduce padding, our tuple packing approach can further improve performance by $10\%$.
Furthermore, by adjusting the limits for edges and nodes, a further $50\%$ improvement has been shown to be possible
on the PCQM4Mv2 molecule graph dataset.
These improvements in reduced padding can benefit any hardware accelerator,
especially when working with static shapes 
and ahead of time compilation strategies.

For datasets with a large variety of potential node-edge combinations,
the complexity of our packing algorithm increases significantly.
Future work is planned to speedup 
the tuple packing algorithm for these kinds of datasets and evaluate the packing approach 
on Graph Neural Networks to determine, how much speedup can be achieved by reducing the padding.

\bibliographystyle{unsrt}
\bibliography{references}

\clearpage

\section*{Appendix}

\begin{lstlisting}[language=Python, caption=Data loading, label=lst:data]
# Copyright (c) 2022 Graphcore
import ogb
from ogb.graphproppred import PygGraphPropPredDataset
from torch_geometric.data import DataLoader
from ogb.utils.mol import smiles2graph
ogb.utils.smiles2graph = smiles2graph
from ogb.lsc.pcqm4mv2_pyg import PygPCQM4Mv2Dataset
import numpy as np
import matplotlib
import matplotlib.pyplot as plt
import os
from collections import defaultdict

# create folder for images
try:
    os.mkdir("graph_packing_tutorial")
except:
    pass

ogb_data = {}

def get_dataset_train(d_name):
    if d_name == "ogbg-pcqm4mv2":
        dataset = PygPCQM4Mv2Dataset(smiles2graph=smiles2graph)
    else:
        dataset = PygGraphPropPredDataset(name=d_name) 
    split_idx = dataset.get_idx_split() 
    train_loader = DataLoader(dataset[split_idx["train"]], batch_size=1, shuffle=False)
    return train_loader

for key in ["ogbg-molhiv", "ogbg-molpcba", "ogbg-code2", "ogbg-pcqm4mv2", "ogbg-ppa"]:
    print("loading:", key)
    ogb_data[key] = get_dataset_train(key)

def get_histogram(data_loader):
    histogram = defaultdict(int)
    for item in data_loader:
        histogram[(item.num_nodes, item.num_edges)] += 1
    return histogram

ogb_histograms = {}
for key in ogb_data:
    print("creating histogram:", key)
    ogb_histograms[key] = get_histogram(ogb_data[key])
    print(ogb_histograms[key])

def get_max_tuples_length(histogram):
    """Get the maximum entry size for each tuple component"""
    maximum_length = []
    for key in histogram:
        if not maximum_length:
            maximum_length = list(key)
        for index, entry in enumerate(maximum_length):
            maximum_length[index] = max(entry, key[index])
    return maximum_length

# getting  max_tuples_length
ogbg_mtl_dict = {}
for key in ogb_histograms:
    ogbg_mtl_dict[key] = get_max_tuples_length(ogb_histograms[key])
print("Max Tuple length:", ogbg_mtl_dict)

def visualize_2D_histogram(histogram, key, dropout=0.01):
    total_count = sum([histogram[(nodes, edges)] for nodes, edges in histogram])
    threshold = total_count * dropout / 100
    num_nodes = [nodes for nodes, edges in histogram if histogram[(nodes, edges)] >= threshold]
    num_edges = [edges for nodes, edges in histogram if histogram[(nodes, edges)] >= threshold]
    image = np.zeros([max(num_nodes)+1, max(num_edges)+1])
    exceptions = []
    for nodes, edges in histogram:
        try:
            image[nodes][edges] = histogram[(nodes, edges)]
        except IndexError:
            exceptions.append((nodes, edges, histogram[(nodes, edges)]))
    if exceptions:
        print("Not visualised:", sum([i[2] for i in exceptions])/total_count*100, "%")
    fig = plt.figure(dpi=200)
    im = plt.imshow(image, cmap=plt.cm.hot_r)
    cb=plt.colorbar(shrink=0.5)
    cb.set_label("Number of samples")
    plt.xlabel("Number of Edges")
    plt.ylabel("Number of Nodes")
    plt.title("Dataset: " + key)
    fig.savefig("graph_packing_tutorial"+os.sep+key+"_dual_histogram.png", bbox_inches="tight")

for key in ogb_histograms:
    visualize_2D_histogram(ogb_histograms[key], key, dropout=0.05)
\end{lstlisting}

\newpage
\begin{lstlisting}[language=Python, caption=Dual Longest-Pack-First Histogram-Packing, label=lst:tuplepack]
# Copyright (c) 2022 Graphcore
def pack_using_dlpfhp(
    histogram,
    max_tuple_length,
    max_tuples_per_pack,
    heuristic=lambda x,y: int(x*y),
    verbose=True,
):
    """Dual Longest-pack-first histogram-packing algorithm.

    Arguments:
        histogram Dict[Tuple[int, int], int]: The histogram of the dataset, it maps
            pairs of node and edge numbers to the number of graphs which match this 
            specific size in the dataset.
        max_tuple_length (Tuple[int, int]): A pair that describes the maximum size of
            the container for each component that must be filled with
            packing. In this example this is a maximum number of nodes or edges.
        max_tuples_per_pack (int | Literal["max"]): This integer parameter limits how
                many tuples/graphs can be combined. If using "max", no limit on packs is
                set, which in some cases can slow down the packing algorithm drastically.
        heuristic (Callable[int, int]): A function which calculates the priority heuristic
            from the histogram key.
    """
    # heuristic assignment
    heuristic_data_list = [(heuristic(a,b), a, b, histogram[(a,b)])
                           for a, b in histogram]
    heuristic_data_list.sort()
    heuristic_data_list.reverse()
    data_list = heuristic_data_list
    max_a, max_b = max_tuple_length[0], max_tuple_length[1]
    max_size = heuristic(max_a, max_b)
    if max_tuples_per_pack == "max":
        max_tuples_per_pack = min(max_tuple_length)
    # Initialize main strategy data dictionary.
    # The key indicates how much space is left.
    # The value is a list of tuples, consisting of counts and respective packs/tuples.
    tmp_strategies_per_length = defaultdict(list)
    strategies_per_length = defaultdict(list)
    for i in range(len(data_list)):  # distribute each bin of histogram
        size, len_a, len_b, n_sequences_to_bin = data_list[i]
        left_size = heuristic(max_a - len_a, max_b - len_b)
        offset = 0 # smallest possible offset for perfect fit
        while n_sequences_to_bin > 0:
            keys = [key for key in tmp_strategies_per_length if key >= size+offset]
            if not keys:
                offset = max_size + 1
            else:
                offset = min(keys)-size
            if (size + offset) in tmp_strategies_per_length:
                for i in range(len(tmp_strategies_per_length[size + offset])):
                    lens_a, lens_b, n_sequences_to_pack = tmp_strategies_per_length[size + offset][i]
                    if (len_a + sum(lens_a)) <= max_a and (len_b + sum(lens_b)) <= max_b:
                        tmp_strategies_per_length[size + offset].pop(i)
                        new_lens_a = lens_a.copy()
                        new_lens_a.append(len_a)
                        new_lens_b = lens_b.copy()
                        new_lens_b.append(len_b)
                        new_size = heuristic(max_a - sum(new_lens_a), max_b - sum(new_lens_b))
                        new_count = min(n_sequences_to_pack, n_sequences_to_bin)
                        # adjust strategies
                        if n_sequences_to_pack > new_count:
                            tmp_strategies_per_length[size + offset].append((lens_a, lens_b, n_sequences_to_pack-new_count))
                        if not tmp_strategies_per_length[size + offset]:
                            tmp_strategies_per_length.pop(size + offset)
                        if new_size == 0 or max_tuples_per_pack == len(new_lens_a):
                            strategies_per_length[0].append((new_lens_a, new_lens_b, new_count))
                        else:
                            tmp_strategies_per_length[new_size].append((new_lens_a, new_lens_b, new_count))
                        n_sequences_to_bin -= new_count
                        offset = 0
                        break
            offset += 1
            if offset + size > max_size:
                new_size = heuristic(max_a - len_a, max_b - len_b)    
                if new_size == 0 or max_tuples_per_pack == 1:
                    strategies_per_length[new_size].append(([len_a], [len_b], n_sequences_to_bin))
                else:
                    tmp_strategies_per_length[new_size].append(([len_a], [len_b], n_sequences_to_bin))
                n_sequences_to_bin = 0
                break
    # merge all strategies
    for key in tmp_strategies_per_length:
        strategies_per_length[key].extend(tmp_strategies_per_length[key])
    # flatten strategies dictionary
    strategy_set = []
    strategy_repeat_count = []
    sum_lens_a, sum_lens_b = [], []
    for key in strategies_per_length:
        for lens_a, lens_b, n_sequences_to_pack in strategies_per_length[key]:
            strategy_set.append((lens_a, lens_b))
            strategy_repeat_count.append(n_sequences_to_pack)
            sum_lens_a.append(sum(lens_a))
            sum_lens_b.append(sum(lens_b))
    if not (max_a == max(sum_lens_a) and max_b == max(sum_lens_b)):
        if verbose:
            print("max discrepancy, reducing sequence length", max_a, max(sum_lens_a), max_b, max(sum_lens_b))
        max_a, max_b = max(sum_lens_a), max(sum_lens_b)
    # efficiency calculation
    empty_tokens_a = int(sum([
        count*(max_a-sum(pack_a)) for count, (pack_a, pack_b) in
        zip(strategy_repeat_count, strategy_set)]))
    empty_tokens_b = int(sum([
        count*(max_b-sum(pack_b)) for count, (pack_a, pack_b) in
        zip(strategy_repeat_count, strategy_set)]))
    packs = int(sum(strategy_repeat_count))
    total_tokens_a, total_tokens_b = int(max_a * packs), int(max_b * packs)
    token_efficiency = (100 - empty_tokens_a / total_tokens_a * 100, 100 - empty_tokens_b / total_tokens_b * 100)
    return strategy_set, np.array(strategy_repeat_count), token_efficiency
\end{lstlisting}


\end{document}